\documentclass[letterpaper, 10 pt, conference]{ieeeconf}
\IEEEoverridecommandlockouts
\usepackage[T1]{fontenc}
\usepackage[latin9]{inputenc}
\usepackage{amsmath}
\usepackage{amssymb}
\usepackage{graphicx}

\usepackage{amsthm}

\newtheorem{thm}{\protect\theoremname}
\newtheorem{defn}[thm]{\protect\definitionname}

\newtheorem{rem}[thm]{\protect\remarkname}

\providecommand{\definitionname}{Definition}
\providecommand{\propositionname}{Proposition}
\providecommand{\remarkname}{Remark}
\providecommand{\theoremname}{Theorem}
\providecommand{\lemmaname}{Lemma}

\title{Optimal Navigation Functions for Nonlinear Stochastic Systems}
\author{Matanya B. Horowitz, Joel W. Burdick
\thanks{Matanya Horowitz and Joel Burdick are with the 
	Department of Control and Dynamical Systems,
        Caltech, 1200 E California Blvd., Pasadena, CA.
        The corresponding author is available at {\tt\small mhorowit@caltech.edu}}%
\thanks{Matanya Horowitz is supported by a NSF Graduate Research Fellowship.%
}
}

\begin{document}
\maketitle

\begin{abstract}
This paper presents a new methodology to craft navigation functions for nonlinear systems with stochastic uncertainty. The method relies on the transformation of the Hamilton-Jacobi-Bellman (HJB) equation into a linear partial differential equation. This approach allows for optimality criteria to be incorporated into the navigation function, and generalizes several existing results in navigation functions. It is shown that the HJB and that existing navigation functions in the literature sit on ends of a spectrum of optimization problems, upon which tradeoffs may be made in problem complexity. In particular, it is shown that under certain criteria the optimal navigation function is related to Laplace's equation, previously used in the literature, through an exponential transform.  Further, analytical solutions to the HJB are available in simplified domains, yielding guidance towards optimality for approximation schemes. Examples are used to illustrate the role that noise, and optimality can potentially play in navigation system design.
\end{abstract}

\section{Introduction}\label{sec:intro}

This paper presents a new method to construct navigation functions that bring a robot operating
under stochastic uncertainty in its control inputs to a desired configuration while avoiding
collision with obstacles.  This is done through analysis of the Hamilton Jacobi Bellman Equation
(HJB) for nonlinear stochastic optimal control problems.  It has recently been found that under mild
conditions the HJB is related through a logarithmic transform to a linear partial differential
equation (PDE) \cite{Theodorou:2011uz,Todorov:2009tha,Kappen:2005kb}.  This paper specializes this analysis to robot navigation functions,
resulting in new planning methods for robots with stochastic nonlinear dynamics.

{\bf Related Work.} This paper touches upon two classical problems: navigations functions (see Section
\ref{sec:review} for a review), and stochastic optimal control problems.  Navigation functions were
introduced by Koditschek and Rimon \cite{Koditschek:1987kt,Koditschek:1990dn,Rimon:1992ch} to remedy the local minima problem in the classical
potential field method of robot motion planning \cite{Khatib:1985hf}, and their early work on
navigation functions focused primarily on the existence and discovery of potential functions whose
gradient would lead a point mass model of a robot from any point in the robot's configuration space
to a desired goal.  Later work extended the navigation function concept to incorporate multiple
agents \cite{howard2002mobile}, and sensory input \cite{lumelsky1990incorporating}.

However, little consideration has been given to including notions of optimality into a navigation
function.  This paper shows how to directly incorporate optimality criteria that can be modeled as
the sum of a (possibly) nonlinear state dependent cost and a quadratic control cost.  This contrasts with the existing focus of the navigation approach, which implicitly defines a decomposition of the the problem into a
trajectory generation method (the solution of the navigation function) and a local feedback
trajectory following control method. This may lead to suboptimality in the path planning or control law, or even instability \cite{Koren:1991hq}. By formulating the problem relative to the Hamilton Jacobi Bellman problem, our method allows the impact of the dynamic model to be directly incorporated into the navigation function, if desired.

 Specifically, we examine the presence or absence both dynamics and  a state dependent cost function in an optimal control problem. We find that if these are neglected, methods similar to those used to generate navigation functions that exist in the literature are recovered \cite{Kim:1992du, loizou2011closed}. The result is that a spectrum of problems, ranging from the full HJB to the classical potential-based navigation function, are made explicit and the tradeoffs in modeling complexity becomes visible. The analysis we present makes it apparent how dynamics may then be incorporated to a navigation function if desired. To our knowledge, this paper also represents the first attempt to formally include stochastic uncertainty into the construction of the navigation function.

Solutions to stochastic optimal control problems pose inherently difficult computational problems.
Classical work on addressing these problems have pursued several approaches.  Systems with linear
dynamics perturbed by Gaussian noise have well known closed form solutions \cite{zhou1996robust}.  For
systems which do not fit these assumptions, several approximation methods, such as receding horizon
control \cite{jadbabaie2001unconstrained} or roll-out methods \cite{bertsekas1995dynamic} have proven to be very useful. Another direction
to solving these problems has been through discretization of the system's state space, leading to a
Markov Decision Process. When framed as such, it is possible to solve the problem exactly
and globally through techniques such as value iteration or linear programming \cite{bertsekas1995dynamic}. If a measure
of error is acceptable, these methods have been made applicable to arbitrarily large state spaces
through Approximate Dynamic Programming \cite{de2003linear}.

This paper follows an alternative line of work--{\em linearly solvable} stochastic optimal control (SOC) algorithms.  The study of linearly solvable SOC problems has recently been studied from two avenues. Work begun by Kappen \cite{Kappen:2005bn} examined the HJB and found that, using a transformation borrowed from quantum mechanics and assumptions on stochasticity, it was possible to find a linear partial
differential equation (PDE) whose solution is the SOC solution. Independently, Todorov \cite{Todorov:2007tq} found an alternate model for control in Markov Decision Processes that allowed for optimal MDP policies to be found as a solution to a set of linear equations, as compared to value iteration.  It was subsequently found that by taking the continuous limit of the discrete Markov model, this same linear HJB PDE is obtained. The HJB can then be understood as the limit process of an MDP. 

Recent research along these lines  has tended towards developing sampling based approaches for solving the resulting linear PDE.  This is done through the use of the Feynman-Kac Lemma, which allows for a linear PDE to be solved by examining the diffusion of a stochastic process. This has been further developed by Theodorou et al. \cite{Theodorou:2009ut, Theodorou:2011uz} into the Path Integral framework. Other research into the study of the computational benefits, and analytic properties, of the linear HJB are available in \cite{Todorov:2009wja, horowitz2014efficient}. Our work begins from the same point, that of Stochastic Optimal Control. However, whereas \cite{Theodorou:2011uz} uses the linearity of the HJB to develop a novel reinforcement learning algorithm with emphasis on numerical solution methodologies, the focus of the present paper is on the application of this flavor of Stochastic Optimal Control to the problem of navigation functions. Connections between these two research fields are emphasized over the specifics of the numerical calculation of the PDE solution, for which we discuss available methods from the literature.

\section{Review: Navigation Functions}\label{sec:review}

Let $\mathcal{CS}$ denote the robot's \emph{configuration space} (or {\em c-space})-- the possible configurations that a robot can occupy.  As is standard, let the subset of $\mathcal{CS}$ where the robot collides with an obstacle define the set of \emph{configuration-space obstacles}, $\mathcal{CO}$, while the {\em free configuration space}, $\mathcal{F}\subset\mathcal{CS}$, is the complement of $\mathcal{CO}$ in $\mathcal{CS}$.  Under the assumption of perfect sensory information, and traditionally also perfect actuation control, the motion planing task is to move the robot from its starting configuration, $q_{init}\ \in \mathcal{F}$ to a desired goal position $q_{d}\ \in\ \mathcal{F}$. One approach to solve this problem is to construct a \emph{navigation
function}: 

\begin{defn} (From \cite{Koditschek:1990dn}) \label{def:nav_def}
Let $q_{d}$ be a goal configuration in $\mathcal{F}$, the free c-space. A map
$\varphi: \mathcal{F} \to [0,1]$ is a \emph{navigation function} if it is

1) smooth on $\mathcal{F}$ (at least a $C^{(2)}$ function);

2) polar at $q_{d}$, i.e., has a unique minimum at $q_{d}$ on the path-connected component of 
$\mathcal{F}$ containing $q_{d}$;

3) admissable on $\mathcal{F}$, i.e., uniformly maximal on the boundary of $\mathcal{F}$;

4) a Morse function.
\end{defn}

Given a navigation function, $\varphi(q)$, the robot's path to the goal from any staring
configuration in $\mathcal{F}$ can be realized by following the gradient $\nabla\varphi(q)$ at each $q$.  The definition assures that the robot will achieve the goal while remaining in $\mathcal{F}$, and not become trapped in a local minima of $\varphi(q)$.

Navigation functions may be constructed in several forms.  In the well-known classical approach of \cite{Rimon:1992ch}, a navigation function may be calculated analytically when the when the bounded problem domain, the obstacle shapes, and the goal region are all diffeomorphic to spheres. Similarly, if the boundary, obstacles, and goal region are star-shaped sets (which are homeomorphic to spheres), then one can compute the navigation function by transforming the problem to a sphereworld, find the sphereworld navigation function, and transforming the function back to the original problem domain.

\subsection{Navigation Function Considerations}

Navigation functions have been successful in part due to their rapid computability and its transparent nature. Furthermore, the navigation function provides both a global plan, as well as a feedback controller that follows the gradient of the navigation function. This allows the total motion planning and execution problem to be abstracted into
a trajectory planner (the path followed by the gradient of the navigation function) and a path following controller \cite{Rimon:1992ch}.  The price to pay for this convenience is optimality: ignoring dynamics in the quest to follow the gradient will rarely result in a procedure that minimizes control effort.  This is in part due to the fact that the system dynamics do not enter into the navigation functions calculation, and may result in unexpected and unstable behavior in some contexts \cite{Koren:1991hq}. The work presented below shows the connection between navigation functions and more general optimal control theory, allowing for system dynamics to be included if this is desirable. It also becomes possible to include more sophisticated weighting of various goals, such as the desire for minimum-time trajectories.

In the construction and intuition behind navigation functions, implicit is the idea of
robustness. Since navigation functions are defined over the entire free configuration space, small deviations from the desired path place the robot in nearby locations where the desirable behavior is similar. Indeed, smoothness of the solution is typically enforced. This work furthers the understanding of robustness properties of navigation functions which has largely heretofore been
only analyzed ad-hoc.

Finally, some approaches for finding navigation functions require difficult calculations and may not extend to complex obstacle geometries. 

\section{A Linear Hamilton Jacobi Bellman PDE} \label{sec:HJB}

We begin by reviewing results in optimal control theory that lead to the construction of the
HJB. This paper considers nonlinear systems that evolve with the following nonlinear stochastic
dynamics
  \begin{equation}\label{eq:stochastic-dynamics}
     dx = \left(f(x)+G(x)u\right)dt+B(x)L\, d\omega
  \end{equation}

where $x_{t}\in\mathbb{R}^{n}$ is the system state at time $t$, $u_{t}\in\mathbb{R}^{m}$ denotes the control input, $\omega$ is a unit variance Brownian motion stochastic process, and $L$ is a state independent noise scaling factor.  The functions $f(x),G(x),B(x)$ are assumed to be smooth  functions of the state and may be nonlinear. We assume that the problem accrues costs $r$ according to
  \begin{equation}\label{eq:cost}
	r(x,u) = q(x) + \frac{1}{2} u^{T} R u
  \end{equation}
where $q(x)$ is any (potentially nonlinear) non-negative function of state. The model for cost is limited to a form quadratic in the controls in order to simplify the HJB equation in a future step of the analysis. The goal is to minimize the following expected cost functional
  \begin{equation}\label{eq:cost-functional}
     J(x_{0:T},u_{0:T})\ =\ \mathbb{E}\bigg[ \phi(x_{T})+\int_{0}^{T}r(x_{t},u_{t})dt\bigg]
  \end{equation} where $\phi$ represents a state-dependent terminal cost, $T$ is the final time of the trajectory,
$\mathbb{E}[\cdot]$ is the expectation operators, and the symbols $x_{0:T}$ and $u_{0:T}$ denote the state and control
over the interval $[0,T]$.

We consider the first-exit problem, wherein the state of the system exists in a compact domain $\Omega$. The system continues to operate, and accrue cost, until it reaches the boundary, $\partial\Omega$, of the
domain at time $T$ whereupon the terminal cost $\phi(x(T))$ is accrued. In the navigation problem, this boundary consists of goals and obstacles in the robot workspace.

A common construction in the optimization literature is the value function, $V(x_t)$, which captures
the ``cost-to-go'' from a given state.  The optimal action follows the gradient of the value
function, bringing the agent into the states with lower cost over the remaining time horizon. The
solution to the optimization problem is, beginning from an initial point $x_{t}$ at time $t$,
  \begin{equation}\label{eq:value-def}
    V\left(x_{t}\right)=\min_{u(\cdot)}\mathbb{E}\left[J\left(x_{t}\right)\right]
    \end{equation}

Using a dynamic programming argument it is possible to derive the Hamilton-Jacobi-Bellman (HJB) equation associated with this problem \cite{fleming2006controlled}, which is found to be
  \begin{equation} \label{eq:original-hjb}
     0=\min_{u}\left(r+\left(\nabla_{x}V\right)^{T}f + 
        \frac{1}{2}Tr\left(\left(\nabla_{xx}V\right)G\Sigma_{\epsilon}G^{T}\right)\right)
  \end{equation}
where $\Sigma_{\epsilon}=LL^{T}$ and the dependency on state is suppressed for brevity.  Since the
control effort enters quadratically into the cost, the optimal control takes the form:
  \[ u^{*}=-R^{-1}G^{T}\left(\nabla_{x}V\right) \]
Substituting the optimal $u$ into (\ref{eq:original-hjb}) yields the following nonlinear, second
order PDE in the cost-to-go $V(\cdot)$:
  \begin{eqnarray*}
       0 & = & q+\left(\nabla_{x}V\right)^{T}f-\frac{1}{2}\left(\nabla_{x}V\right)^{T}
         G R^{-1} G^{T} \left(\nabla_{x} V \right)\\
            & + & \frac{1}{2} Tr \left(\left(\nabla_{xx}V\right) B \Sigma_{\epsilon} B^{T}\right)
  \end{eqnarray*}
The difficulty of solving this nonlinear, second order PDE often prevents practitioners of optimal
control from attempting to solve for the value function directly. However, it has recently been
found \cite{Kappen:2005bn,Todorov:2009wja,Theodorou:2011uz} that with the assumption
  \begin{equation}\label{eq:noise-assumption}
     \lambda G(x)R^{-1}G(x)^{T}=B(x)\Sigma_{\epsilon}B(x)^{T}\triangleq\Sigma_{t}
    \end{equation}
and the logarithmic transformation 
   \begin{equation}\label{eq:log-transform}
     V=-\lambda\log\Psi
   \end{equation}
one can obtain, after substitution and simplification, the following {\em linear} PDE 
   \begin{equation}\label{eq:hjb-pde}
       0 = -\frac{1}{\lambda}q\Psi + f^{T}(\nabla_{x}\Psi) + 
           \frac{1}{2} Tr \left(\left(\nabla_{xx}\Psi\right) \Sigma_{t}\right)\ .
\end{equation}
This transformation of the value function, which we call here the \emph{desirability}, provides an additional, computationally appealing, method through which to calculate the value function.  The
solution to the desirability may readily be transformed by (\ref{eq:log-transform}) to obtain the value function, which may then be used for execution.  Note that condition (\ref{eq:noise-assumption})
can roughly be interpreted as a controllability-type condition: the system must have sufficient control to span (or counterbalance) the effects of input noise on the system dynamics. Furthermore, it must be ``cheap'' for the system to push in directions where noise is high, and expensive were noise is low. Additional discussion may be found in \cite{Todorov:2009wja}.

Note that Eq. \eqref{eq:hjb-pde} is in particular an \emph{elliptic} PDE, and therefore obeys the maximum principle for elliptic PDEs \cite{evans1998partial}. This implies that there exist no local minima or maxima in the interior $\Omega$ of these HJB solutions, satisfying the Morse property of navigation functions in Definition \ref{def:nav_def}.

\section{Navigation Functions through Optimal Control}

This section will first reduce the SOC problem introduced above to the standard setting of navigation functions by sequentially incorporating the assumptions which hold in the classical navigation function setting.  These successive eliminations of terms will then illuminate some connections between our approach and classical navigation function approaches.  Finally, we will suggest how an approximate minimum time problem can be formulated in this approach.

\subsection{Reduction to the Navigation Function}

{\bf Dynamics.} Since the classical navigation function approach implicitly decouples the  trajectory generation problem from the trajectory following control design, the dynamics of a  specific system are ignored.  Hence, in our parallel development of the Navigation HJB equation, the dynamic term may be dropped $f:=0$. This results in the \emph{Navigation PDE}:
  \[ 0=-\frac{1}{\lambda}q\Psi+\frac{1}{2}Tr\left(\left(\nabla_{xx}\Psi\right)
           \Sigma_{\epsilon}\right). \]
Similarly, the classical navigation function setting does not consider spatially dependent costs. Thus, the state-dependent term in the cost function, $q(x)$, may be simplified to a free scalar parameter $q:=\alpha$, producing the PDE
  \begin{equation}\label{eq:screened-nav}
    0=-\frac{\alpha}{\lambda}\Psi+\frac{1}{2}Tr\left(\left(\nabla_{xx}\Psi\right)
       \Sigma_{\epsilon}\right)
\end{equation}
We will term this PDE as the \emph{Augmented Navigation PDE}, as it incorporates additional cost information as compared to traditional navigation functions, but does not include the effects of system dynamics. The effect is that those states that appear only in the dynamics, and are not the workspace states, may be neglected as well. In reverse, if one wishes to include dynamics, their presence in $f(x)$ will require the additional of these states as dimensions in the HJB PDE.

Interestingly, this PDE is well known as the homogeneous \emph{Screened Poisson Equation}, and has found applications in image processing \cite{bhat2008fourier}. Of interest here are the observations that this is a second order PDE with isotropic diffusion and mass terms, a situation which has been well studied \cite{pozrikidis2005introduction}.  


{\bf Boundary Costs.} The boundary conditions for the PDE correspond to the penalty accrued as the robot exits the configuration domain and collides with an obstacle or reaches the goal state.  In (\ref{eq:cost-functional}) this is represented as the terminal cost $\phi$. Recall that by (\ref{eq:log-transform}), this terminal cost must be transformed along with the value function. Thus, we have for the boundary condition
   \begin{equation}
	\Psi\mid_{\partial\Omega}=e^{-\frac{\phi}{\lambda}}\label{eq:screened-boundary}
   \end{equation}
where $\partial \Omega$ is the boundary of the operating domain, $\Omega$.  Classically, the cost assigned to a collision has been modeled as uniform over all obstacles (Property 3 of Definition \ref{def:nav_def}), and we may thus set $\phi(x_{T})=c$ for an arbitrary constant $c$.  Other choices are certainly possible.

\begin{rem}
We note that the boundary conditions of a linear PDE, as we have here, obey the principle of superposition. Solutions to problems may be composed at essentially zero cost, a potentially significant savings for systems that require many plans over a common domain, or require a change in the workspace. This topic is explored in \cite{Todorov:2009tha} and \cite{HorowitzComp}.
\end{rem}

The free variables $q(x)$ and $R$ define a notion of cost, and therefore a notion of optimality. The inclusion of these variables allows us to compare navigation functions according to their perceived cost, and furthermore to declare navigation functions optimal with respect to a choice of criteria. Such criteria has traditionally been eschewed in favor of simplicity in construction of the navigation function, and hence our framework may be said to be a slight generalization, bringing
notions of optimality into consideration.

{\bf Control-dependent costs.} Recall that our initial definition of cost (\ref{eq:cost}) includes a control dependent term.   Navigation functions have traditionally been unconcerned with the control effort. Recall that the assumption on control effort and noise (\ref{eq:noise-assumption}) needed to realize a linearly solvable HJB PDE is:
   \begin{equation}\label{eq:noise-lambda-constraint}
 \lambda G(x)R^{-1}G(x)^{T} = \Sigma_{t} 
\end{equation}
where $\Sigma_{t}$ is fixed as a function of the known control vector field matrix, $G(x)$, and noise characteristics, $B(x)$ and $\Sigma_{\epsilon}$. The control effort penalty $R$ cannot be brought to zero naively without violating this assumption.  It is possible to compensate for this limitation by using the free parameter $\lambda$ to maintain the underlying relation in this assumption. That is, set $\lambda=\beta$ and define $R=\beta\tilde{R}$, yielding expressions
  \begin{eqnarray}
    \lambda G(x)\left(\beta\tilde{R}\right)^{-1}G(x)^{T} & = & \Sigma_{t} \label{eq:beta-lambda}\\
      G(x)\tilde{R}^{-1}G(x)^{T} & = & \Sigma_{t} \nonumber
  \end{eqnarray}
which is independent of $\beta$, allowing the control penalty cost to be reduced to zero. The difficulty is that as $\lambda\to0$, (\ref{eq:screened-nav}) becomes nonsensical in the limit. Fortunately, we have assumed no cost over the states and set $\alpha=0$ to produce the {\em Navigation PDE}
  \begin{equation}\label{eq:laplace-nav}
     0=Tr\left(\left(\nabla_{xx}\Psi\right)\Sigma_{t}\right)
  \end{equation}
which is recognized to be Laplace's equation scaled according to the system noise characteristics. The practical cost incurred by this reduction of the complete SOC HJB, Equation (\ref{eq:hjb-pde}), to Equation (\ref{eq:laplace-nav}) is that we have eliminated consideration of control effort and state dependent penalties, which is natural in the robotics setting. Interestingly, Laplace's equation has been used previously in the generation of navigation functions \cite{Connolly:1990tc, Kim:1992du, loizou2011closed}. In this prior work, the authors suggested the use of Laplace's equation, with the motivation that solutions to Laplace's equations can be shown to have no local minima over their domain. The following theorem justifies this from an optimality perspective, albeit through the transformation (\ref{eq:log-transform}).

\begin{thm} \label{thm:Nav-PDE}
The optimal robust desirability function absent costs over state is given by $V=-\lambda \log \Psi$ where $\lambda$ is according to \eqref{eq:noise-lambda-constraint}, and $\Psi$ is the solution to Laplace's equation over the domain $\Omega$:
  \begin{eqnarray*}
    0 & = & Tr\left(\left(\nabla_{xx}\Psi\right)\Sigma_t\right)\\
     \Psi\mid_{\partial\Omega} & = & e^{-\frac{\phi}{\lambda}}
  \end{eqnarray*}
\end{thm}

\begin{rem}
There is an interesting trade-off resulting from Eq. \eqref{eq:noise-lambda-constraint}. Define $ \tilde{\Sigma}_t \triangleq \gamma \Sigma_t $ in order that the noise may be scaled by $\gamma$. Define $\lambda$ = $\beta$, as in \eqref{eq:beta-lambda} in order to scale the control penalty. Then $\lambda = a \beta \gamma$ for some fixed constant $a$. The result is that a scaling of the control effort has the same effect on the solution as a scaling of the noise, and these manifest only
through the transformation \eqref{eq:log-transform} and the boundary conditions, and surprisingly not through the differential constraints on $\Psi$ as $\gamma$ is simply cancelled in Eq.  \eqref{eq:laplace-nav}. What does have an effect, however, is the directional influence of $\Sigma_t$, i.e. the solution will incorporate paths that have beneficial drift.

Due to the exponential dependence of Eq. \eqref{eq:log-transform}, one cannot realize the limit $\beta \to 0$. Instead, $\beta$ is chosen based on magnitude of the noise, or the level of control penalty, depending on the perspective. \label{beta-remark}
\end{rem}

\begin{rem}
Laplace's equation has been justified in the presence of noise here, whereas it was previously justified due to its lack of local minima.  It is notable that in this case the deterministic and stochastic case share their solution, subject to a scaling.
\end{rem}

Two PDEs (\ref{eq:screened-nav}), (\ref{eq:laplace-nav}) have been produced in this initial analysis.  The first of these allows one to naturally incorporate several optimality criteria into the concept of a navigation function, while the second is especially simple and creates a connection to previously inspired navigation function formulae.

\subsection{Approximate Time-Optimal Navigation Functions} 

The freedom afforded by the optimality parameters allows for the solution to be biased towards time-optimal navigation functions by penalizing time spent away from the goal. This is accomplished by setting $q(x)=c$ for some constant $c$. Control effort may also be adjusted through the free parameter $\lambda$ and decreased relative to the state cost. The robustness of the navigation function to noise may also be controlled through the noise characteristics defined by $\Sigma_{\epsilon}$, and again reduced. As the value of the constant $c$ is increased while parameters $\lambda$ and $\Sigma_{\epsilon}$ are increased, cost is accrued only when the system remains outside the goal region.  The optimal action during this time is to take the quickest path to the goal, ignoring the amount of control effort used.

\subsection{Analogy with electro-statics}

It is interesting that early researchers on potential field navigation methods were naturally drawn towards analogies with electro-statics.  Khatib's seminal work \cite{Khatib:1985hf} conceptually frames the collision avoidance problem as a process of adding potential fields that would repulse or attract the point robot mass in much the same way as electrostatic fields might. We now see that this intuitive notion of attractive and repulsive forces can also be grounded in notions of
optimality. Indeed, for the Navigation PDE of Eq. (\ref{eq:laplace-nav}), the analogy is exact in desirability-space, with the representation of obstacles and goals manifesting identically to static charges on their surfaces. However, a logarithmic transform improves the solution from an intuitive one to one which is optimal.

\subsection{Convergence of the solution trajectories}
Due to the presence of the control cost in the solution to the HJB, it isn't possible to directly determine the probability the system successfully reaches a goal region, where failure is dictated by the modeling assumption of stochastic forcing in \eqref{eq:stochastic-dynamics}. As mentioned in Remark \ref{beta-remark}, entirely removing the affect of this cost component isn't possible as the necessary formulae break down in the limit. However, it is possible to take a small value for the control cost $R$. The cost-to-go is then predominantly governed by the cost of colliding with an inadmissible boundary. By setting the boundary conditions for obstacles to $\phi=1$, the cost-to-go then becomes a \emph{conservative} approximation of probability of success in reaching the goal, i.e. the cost-to-go will overstate the probability of failure by also including the cost of future trajectories' control effort in its value. For even moderately small values of $R$, this approximation may not be overly conservative.

It is important to note that of the various PDEs proposed, only \eqref{eq:laplace-nav} guarantees non-collision. Specifically, the penalty accrued on the way to the goal when $q(x)\ne 0$ may be greater than the boundary penalty (for instance if a high time-penalty was applied), and the system may simply plan to hit the nearest obstacle. This may be ameliorated by simply increasing the obstacle penalty to a sufficiently high value, but calculation of such a value a-priori remains an open question.

\subsection{Range of Navigation Functions}
We briefly review the results presented up to this point. Three distinct PDEs have been presented, beginning from the HJB and then constructed by neglecting a term. Our goal was not the construction of the Laplacian, but showing its justification as in fact the optimal solution given the specific assumptions of matched noise \eqref{eq:noise-assumption}, and a neglect of state dependent cost and dynamics. It is also now clear how existing Laplacian techniques may be augmented towards optimality with a simple exponential transformation \eqref{eq:log-transform}, as well as through the inclusion of dynamics to form the Navigation PDE \eqref{eq:hjb-pde}, which may also have the state cost simply set to zero. This informs the discussion as navigation functions are brought to bear in novel domains \cite{loizou2012navigation}.
\section{Numerical Solutions}

The linear HJB is a PDE, and therefore appropriate for conventional numerical tools. These include techniques such as the Finite Difference Method (FDM), the Finite Element Method (FEM), or even the Boundary Element-Fast Multipole Method (FMM) \cite{coifman1993fast, pozrikidis2005introduction}.  In a number of papers Todorov \cite{Todorov:2009tha,Todorov:2009ul,Zhong:2011uv} has looked into a number of alternatives based on the structure of the discretized problem as well. In this work, we use the FDM due to its simplicity as the focus of this work is not on numeric computation.

The mentioned methods are quite fast in practice and may scale up to approximately six dimensions. However, in high dimensions such techniques fall prey to the curse of dimensionality. Recent work using sparse tensor decompositions \cite{Horowitz:hd} have allowed for these HJBs to be solved in dimensions up to twelve, with higher dimensions possible. Alternatively, the Feynman-Kac Lemma may be used to solve the problem through sampling \cite{Theodorou:2011uz,Kappen:2005bn} by simulating a Brownian motion with the PDE related to its generator. Such methods scale independent of dimension (albeit with a constant that is dimension-dependent) but face the drawback that each execution of the algorithm is only able to return the solution to the PDE at an individual point.

\section{Examples}\label{sec:examples}

The approach is also illustrated several numerical examples. Each of these is solved with an $h=0.1$ discretization of the domain and the use of a simple Finite Difference Method.

\subsection{Problems with Simple Solutions}

In simple domains it is possible to find an analytical or simply computed solution to (\ref{eq:screened-nav}), (\ref{eq:laplace-nav}). For instance, suppose a point robot is commanded to move to a goal location located at the origin of a two-dimensional configuration space with no obstacles, while it is perturbed with noise whose characteristics are uniform across the configuration space.  The solution to the associated Navigation PDE (\ref{eq:laplace-nav}) corresponds to the solution of Laplace's equation for a point potential, i.e.  the fundamental solution, which is
   \[ \Psi=-\frac{\log\left(r\right)}{2\pi} \] 
where $r$ is the distance from the robot's current configuration to the origin \cite{evans1998partial}. The solution to the Augmented Navigation PDE (\ref{eq:screened-nav}) for this problem may be found as follows. Taking the Fourier Transform of the equation yields
  \[ \left(k^{2}+\frac{\alpha}{\lambda}\right)\tilde{\Psi}=1 .\]
The fundamental solution is then found by solving for $\tilde{\Psi}$ and then finding the inverse Fourier Transform, which yields
  \[ \Psi=\frac{1}{2\pi}K_{0}\left(\sqrt{\frac{\alpha}{r}}\lambda\right) \]
with $K_{0}$ the modified Bessel function of the second kind. 

\subsection{Effect of Noise on Corridor Navigation}
\begin{figure}
\begin{centering}
\includegraphics[scale=0.25]{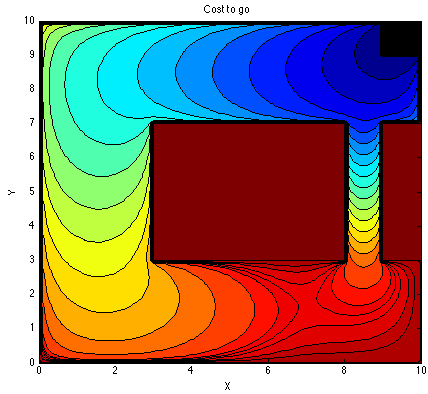}
\includegraphics[scale=0.25]{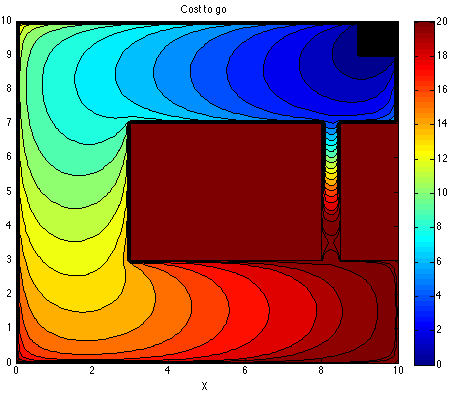}
\par\end{centering}
\caption{Navigation function for a two dimensional point-mass robot calculated according to 
Eq. \eqref{eq:laplace-nav} with varying corridor widths. }
\label{fig:corridor-results}
\end{figure}

Next, we demonstrate the method for a two dimensional robot whose task is to reach the top right corner of a square configuration space. The domain has two obstacles, creating a pair of corridors that the robot must traverse if it begins in lower portion of the configuration space. For this example $\Sigma_t= 2 I_{2\times2}$, $\lambda=1$, and the width of the thinner corridor is set to 1.5 and 2 distance units for comparison. The resulting solutions are shown in Fig. \ref{fig:corridor-results}.
This example shows why it can be important to include noise in the construction of a navigation function. Consider the situation when the robot starts near the bottom of the figure.  In both environments, robot can travel through two different corridors to reach the goal.  In both environments, the navigation functions lack local minima, as expected. In the left-hand environment, the robot can potentially choose between a wide corridor, and a medium width corridor.  The choice of the corridor will depend upon the robot's specific starting configuration, as it is safe to traverse both corridors.  In the right hand figure, the robot can potentially choose between the same large corridor, or a very narrow corridor (whose width is $\frac{3}{4}$ the magnitude of the noise variance).  For almost all starting positions, the navigation function guides the robot away from the potentially dangerous narrow corridor, unless the robot happens to start positioned well into the narrow corridor.  This intuitively logical result occurs because the potential for collision in the narrow corridor places too high a cost on that potential path.

\begin{rem}
The solution to Eqs. \eqref{eq:screened-nav}, \eqref{eq:laplace-nav} take place in exponentiated coordinates, and for many examples tend to be close to zero for large regions of the state space. It is therefore usually more useful to consult the value function directly rather than examining the desirability.
\end{rem}

\subsection{Maze}

\begin{figure}
\begin{tabular}{l l}
\includegraphics[scale=0.25]{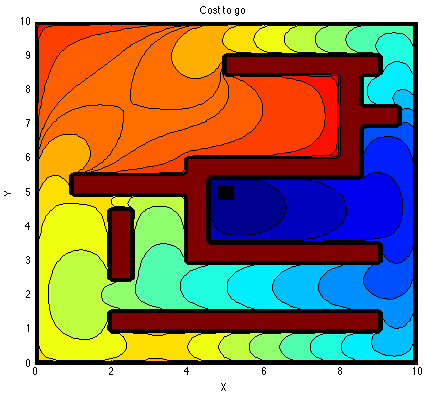} &
\includegraphics[scale=0.25]{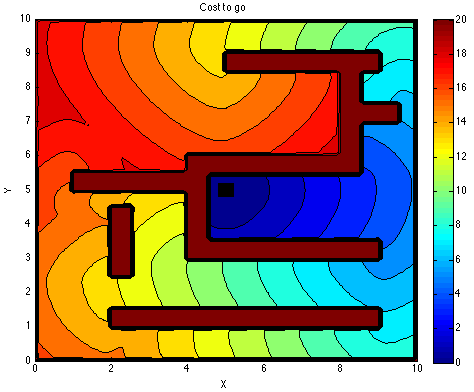} \\
\includegraphics[scale=0.25]{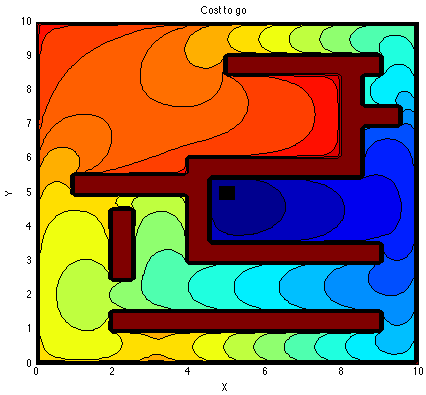} &
\includegraphics[scale=0.25]{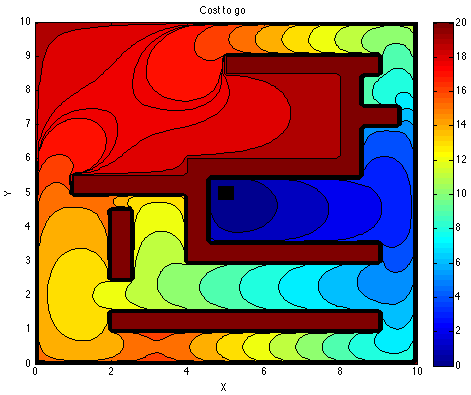} 
\end{tabular}

\caption{Navigation function for a two dimensional point-mass robot in a maze-like environment with equi-dimensional noise. On the top-left is the standard navigation function calculated according to \eqref{eq:laplace-nav}.  On the top-right, the minimum time-criteria is approximated by taking  $\alpha=100$, $\lambda=.04$ in \eqref{eq:screened-nav}. The bottom-left and -right have $\Sigma_t=4 I_2,4.6 I_2$, and $\lambda=.4, .46$ respectively. The obstacles and boundary are chosen to have penalty of 20 units, while the goal region has a penalty of 0 units. The goal is located at the  origin. \label{fig:Navigation-maze}}  
\end{figure}

The second example shows that complicated environments can be well handled by this method, and also highlights the effects of including additional cost criteria into the navigation function. The same robot dynamics and same noise distribution of the previous example are used, however the obstacles are placed in a more complicated maze-like pattern. The Augmented Navigation PDE of Eq. (\ref{eq:screened-nav}) is used in order to incorporate the additional optimality criteria.  The resulting navigation functions are shown in Figure \ref{fig:Navigation-maze} with several noise and cost configurations..

The results compare the use of \eqref{eq:laplace-nav} with the additional cost criteria. It is seen that the solutions of \eqref{eq:laplace-nav} and \eqref{eq:screened-nav} are qualitatively similar. As $\lambda$ is decreased in the general case, this has the interpretation of either increasing the control cost weighting, $R$, or increasing the noise covariance. The solutions do not change qualitatively, only the magnitude of the cost-to-go. In contrast, the approximated minimal-time solution is characterized by shortest-path level sets. These level sets are characterized by straight lines near corners, in contrast to the circular level sets of other examples, as would be expected for a shortest path solution amongst piecewise-linear obstacles.

\begin{rem}
For some choices of costs $q,R$ the navigation function may bring the robot directly into an obstacle. This is no contradiction, as framing the problem through the lens of optimal control allows for freedom on the placement of boundary conditions. The penalty for hitting an obstacle, if chosen improperly, may be less than the cost to traverse the domain and enter the goal region, and thus the most economical choice is for the robot to simply collide with the boundary. This isn't a
problem when using
\eqref{eq:laplace-nav}. 
\end{rem}

\subsection{Grasping}

\begin{figure}
\begin{centering}
\includegraphics[scale=0.35]{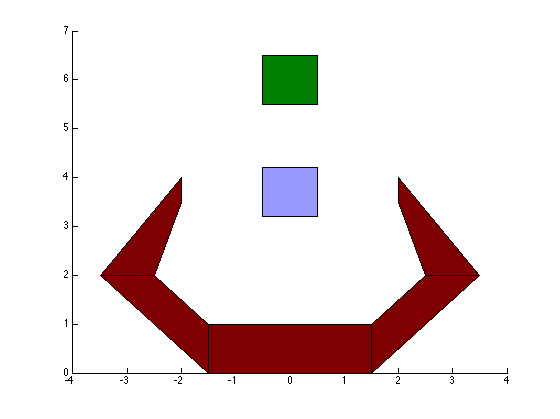}
\par\end{centering}

\caption{An illustration of a nut grasping task. The square nut is shown in green while the gripper is shown in red. The blue region denotes a range of acceptable nut locations relative to the gripper. Each of these colored regions is transformed into configuration space, with the intersection of gripper and nut an obstacle, and intersection between the acceptable locations and the nut a goal. \label{fig:nut-illustration}}
\end{figure}

\begin{figure}
\includegraphics[scale=0.45]{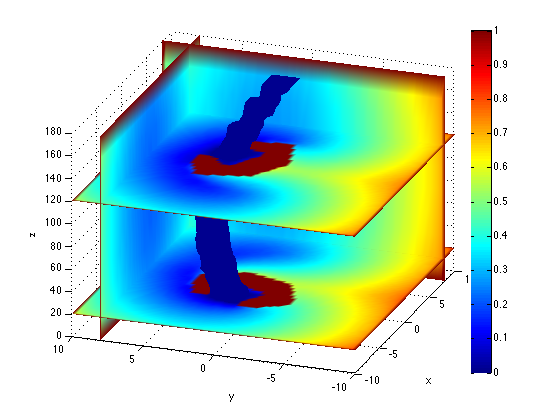}
\caption{Cross sections of navigation function for the nut grasping task illustrated in Figure \ref{fig:nut-illustration}. Parameters used are $\alpha=0.02,R=.02 I_{3\times 3}, \Sigma_t = 5 I_{3 \times 3}$ and the boundary costs are set to $\phi=1$. Spatial discretization in the $x-$ and $y-$coordinates are $h_x = h_y =0.25$ and in the angular direction $h_\theta= 20^{\circ}$. The goal region isosurface is displayed in dark blue.
\label{fig:Results-Nut-Grasping}}

\end{figure}

A final example is of a simple planar grasping task wherein a gripper must be positioned in the  plane so that it may close and grasp a small nut, an illustration of which is shown in Figure \ref{fig:nut-illustration}. The goal of the problem is to move the end effector to one of a continuum of desirable locations surrounding the object in such a way that the gripper's orientation places the jaws around the nut, whereupon a simple closing action will reliably grasp the nut. The problem is transformed into the system's configuration space, and then solved in the Optimal Navigation framework, with the results shown in Figure \ref{fig:Results-Nut-Grasping}. As the method treats the goal states as a set of boundary conditions, the relatively large goal region is handled easily.

The example illustrates the approach for a non-point mass robot, and illustrates the smoothness of the navigation function over the domain in a typical manipulation task. It is easily to see the optimal path of the gripper, wherein it smoothly rotates, following the basin of low cost in blue, until the nut is captured.

\section{Discussion}

This paper introduced a generalization of the navigation function framework to include system noise, dynamics, and cost criteria. Philosophically, this paper links the classical robotics subject of navigation functions with recent advances in Stochastic Optimal Control.  Previous results that were developed on a somewhat ad-hoc basis have been shown to be related to optimality considerations, and the intuition which led to their development well placed. Remarkably, many existing results using harmonic potentials can be shown to be optimal for a particular configuration of noise and cost models when the solution is simply adjusted by a logarithmic transformation.

We also showed how navigation functions can approximately incorporate minimum time task requirements. From a practical point of view, the methods developed here to construct the navigation functions can be applied to environments, obstacles, and robots with arbitrary smooth geometries. Furthermore, it allows for the results of existing methods (i.e. \cite{894676}) to be compared against the underlying \emph{optimal} solution to the problem, if the system's noise characteristics are captured by our model. The ability to solve for the navigation function in terms of a linear PDE also expands the space of algorithms for calculating solutions.

Our numerical experiments show that solutions in configurations space having dimension up to 5 can be computed on a desktop PC computer.  Future work will explore specialized numerical algorithms  in an attempt to expand the practical computational limits of the approach. Additionally, it may be possible to use the results presented here to inform the choice of artificial potential fields, yielding navigation functions that may be assembled quickly, but that better approximate the optimal navigation function for the problem.

\bibliographystyle{abbrv}
\bibliography{navigation_references}

\end{document}